\documentclass{article}

\usepackage{arxiv}

\usepackage[utf8]{inputenc}
\usepackage[T1]{fontenc}
\usepackage{url}
\usepackage{booktabs}
\usepackage{amsfonts}
\usepackage{amsmath}
\usepackage{nicefrac}
\usepackage{microtype}
\usepackage{graphicx}
\usepackage[numbers]{natbib}
\usepackage{float}
\usepackage{array}
\usepackage[colorlinks=true,linkcolor=blue,citecolor=blue,urlcolor=blue]{hyperref}
\usepackage{cleveref}

\title{Tiny-Engram: Trigger-Indexed Concept Tables for Generative Vision}

\date{}

\newif\ifuniqueAffiliation
\uniqueAffiliationtrue

\ifuniqueAffiliation
\author{ Runyuan Cai \quad Yiming Wang \quad Yu Lin \quad Xiaodong Zeng\\
    \vspace{0.5em} \\
    \textit{AutoArk-AI} \\
    \vspace{0.5em} \\
    \texttt{\{runyuan.cai, yiming.wang, yu.lin, xiaodong.zeng\}@autoark.ai} \\
    \url{https://github.com/AutoArk/TinyEngram}
}
\else
\usepackage{authblk}

\author[1]{Runyuan Cai}
\author[1]{Yiming Wang}
\author[1]{Yu Lin}
\author[1]{Xiaodong Zeng}
\affil[1]{AutoArk-AI}
\fi

\renewcommand{\undertitle}{Working Technical Report}
\renewcommand{\shorttitle}{Tiny-Engram}

\hypersetup{
pdftitle={Tiny-Engram: Trigger-Indexed Concept Tables for Generative Vision},
pdfsubject={cs.AI},
pdfauthor={Runyuan Cai, Yiming Wang, Yu Lin, Xiaodong Zeng},
pdfkeywords={Tiny-Engram, TinyEngram, Concept Tables, Generative Vision, Personalization},
}

\newcommand{\method}{Tiny-Engram}

\newcommand{\trigger}{\texttt{Aldric Vortex-9 CyberNebula}}

\begin{document}
\maketitle

\begin{abstract}
Current personalization methods for generative vision models typically encode new concepts through continuous adapters or weight updates, yet provide limited control over whether and when a concept should be retrieved. In this work, we introduce \method{}, a compact trigger-indexed concept table that gives visual memories an explicit lexical address and activation boundary inside frozen image and video generators.

\method{} parameterizes each concept as a small set of memory entries indexed by registered n-gram matches, which modulate text-encoder hidden states only within the matched trigger region. Outside this lexical support, the conditioning pathway is identical to that of the frozen base model. Across both single-encoder latent diffusion and multi-encoder diffusion-transformer backbones, this formulation binds a rare trigger phrase to a target identity while preserving compositional control from the surrounding prompt. We further evaluate the same table-based memory in a text-conditioned video generation setting, where the trigger path reliably alters the generated subject but fine-grained identity persistence across held-out video prompts remains limited.

Taken together, these results suggest that small, explicitly addressed concept tables are a practical route to modular visual personalization, with strongest evidence in image generation. For video diffusion, the remaining gap points to a broader requirement: temporally stable identity likely depends on tighter coupling between text-side memory and the evolving visual state, motivating future work on memory injection beyond the text-conditioning interface.

\end{abstract}


\section{Introduction}

Model adaptation is often constrained by two requirements: the update should be small enough to train and store cheaply, and it should not unnecessarily disturb behavior already present in the pretrained model. Parameter-efficient fine-tuning (PEFT) methods address the first requirement by using adapters, continuous prompts, low-rank updates, or activation scaling \citep{houlsby2019parameter,li2021prefix,lester2021power,hu2022lora,liu2022fewshot}. Retrieval-augmented and memory-augmented language models address a related problem by separating part of model behavior from dense parametric storage \citep{guu2020realm,lewis2020rag,khandelwal2020knnlm,borgeaud2022retro}. Neither direction, by itself, guarantees that an added capability has a discrete activation condition inside the model.

This activation boundary matters for modular adaptation. A continuously active adapter can affect prompts outside the intended domain, and independently trained concept modules can interact when merged or composed. For many personalization and domain-specialization settings, it is useful to ask for stronger locality: the learned update should be applied only when a registered textual condition is present, while all unmatched inputs should follow the original model path.

DeepSeek's Engram work, commonly discussed as a key memory primitive in the DeepSeek-V4 research line, gives one answer to this question. It proposes conditional memory through scalable $N$-gram lookup, so that matched token spans retrieve learned memory entries inside the model \citep{cheng2026conditional}. \method{} operationalizes this principle as a compact adaptation module: a registered $n$-gram in the prompt indexes a learned concept table, and the retrieved vector modifies hidden states only at the matched span. Thus the learned memory is tied to explicit lexical support rather than applied as a globally active adapter.

We first validate TinyEngram in the language-model setting, where Engram was originally proposed. Using Qwen3 as the frozen backbone \citep{yang2025qwen3}, we insert Engram modules into selected transformer layers and train only the added memory parameters. In biomedical adaptation, the Engram branch learns domain-specific behavior while preserving broad MMLU performance \citep{hendrycks2021mmlu}. In a high-intensity function-call adaptation setting, a tuned Engram checkpoint reaches comparable or lower validation loss than a LoRA baseline while retaining stronger TruthfulQA performance \citep{lin2022truthfulqa}. These results establish the premise used in the rest of the paper: trigger-indexed memory can be optimized as a PEFT module, and its localized activation can improve resistance to catastrophic forgetting under the tested adaptation pressure.

This premise motivates the extension to generative vision, where personalization requires binding a textual trigger to a visual concept without changing unrelated prompts. Existing text-to-image personalization methods bind new concepts through learned tokens, subject-specific fine-tuning, attention updates, or image-prompt adapters \citep{gal2022textual,ruiz2023dreambooth,kumari2023custom,wei2023elite,ye2023ipadapter}. In latent diffusion and diffusion-transformer models, text encoders provide the conditioning states that guide image synthesis \citep{radford2021clip,raffel2020t5,rombach2022latent,esser2024scaling}. This makes the text-conditioning stream a suitable interface for trigger-indexed memory: TinyEngram can modify the prompt representation while leaving the image backbone frozen. In SD1.5 \citep{rombach2022latent}, it injects Engram vectors through the single CLIP text encoder \citep{radford2021clip}. In SD3.5 \citep{esser2024scaling}, it extends the same mechanism to the CLIP-L, OpenCLIP-G, and T5 encoder stack \citep{radford2021clip,ilharco2021openclip,raffel2020t5}. This multi-encoder setting introduces a practical scale issue: the injected memory must have comparable effect across text streams with different hidden-state magnitudes. We address this with a relative-norm injection rule. Across the image experiments, a rare textual trigger is bound to a target visual identity while the U-Net or DiT-style denoising backbone remains frozen \citep{ronneberger2015unet,peebles2023dit,esser2024scaling}, and no-trigger controls preserve the base generation path.

We also include a preliminary video extension. Video diffusion models add temporal and motion constraints to the image-generation problem, making identity personalization more demanding than in single-frame synthesis \citep{wu2023tuneavideo,guo2024animatediff,chen2023videocrafter,wan2025}. On Wan2.2 TI2V-5B, text-side Engram injection clearly changes generation, but does not yet match the identity stability observed in SD3.5 images. We use this result as a boundary case rather than as the main evidence for the method.

This report makes four contributions: a trigger-indexed Engram formulation for parameter-efficient adaptation; a text-encoder memory-injection method for image personalization in frozen generative backbones; a relative-norm injection rule for calibrated text-stream updates; and an initial video study that clarifies both the promise and current limitation of text-side memory beyond images.

\section{Method}

\subsection{Overview}

\method{} is a trigger-indexed PEFT module. At the most general level, it adds trainable memory residuals to selected hidden states of a frozen backbone, and activates those residuals only when the input contains registered or hashed $n$-gram evidence. The base model parameters remain fixed; only the Engram memory parameters and lightweight gates or scales are optimized. This gives the adaptation both small storage cost and an explicit activation boundary.

The visual method studied in this paper is a specialization of this formulation for text-conditioned image and video generators. Let $G$ denote an image or video diffusion model with one or more text encoders $\{E_m\}_{m=1}^M$. The generator backbone, variational autoencoder, sampler, and original text-encoder weights are kept fixed. For each wrapped text encoder, \method{} adds only two trainable objects: a table of concept vectors and a per-channel injection-scale vector. A prompt that does not contain the registered trigger follows the frozen model path; a prompt that contains the trigger receives a residual update only on the matched trigger tokens.

We instantiate the visual experiments as a controlled study of binding one source appearance to a new fictional trigger name. The source visual concept is the protagonist appearance from Hideo Kojima's \emph{Death Stranding}, while the prompt-side address is the fictional phrase \trigger{}. This name is chosen to make the binding test conservative. It is not a real character or common entity name, so the frozen generator should have no useful identity prior to retrieve. At the same time, its components---``Vortex'', ``CyberNebula'', and the model-like suffix ``9''---carry a science-fiction and space-oriented lexical bias that tends to elicit cybernetic or interstellar subjects from the base generator, rather than the grounded human protagonist appearance used as supervision. The experiment therefore asks the Engram table to bind a deliberately misleading lexical address to the reference appearance. A plain personal name would weaken this control, because existing name priors or accidental concept collisions could partially explain the result.

\subsection{General Engram PEFT Formulation}

For a frozen backbone $F_\theta$, let $H^r \in \mathbb{R}^{B \times L \times d}$ denote the hidden states at an adapted site $r$, such as a transformer layer in a language model or a text-encoder output in a generative vision model. An Engram module $A^r_\phi$ computes a residual from the input token sequence and the current hidden states,
\[
    \widetilde{H}^r = H^r + A^r_\phi(x,H^r),
    \qquad
    \theta \ \text{frozen}, \quad \phi \ \text{trainable}.
\]
The input sequence is decomposed into $n$-grams. In the language-model implementation used for the supporting Qwen experiments, these $n$-grams are compressed, hashed into layer-specific multi-head memory tables, projected back to the backbone hidden width, and gated by the local hidden state before being added as a residual. Engram modules are inserted only at selected decoder layers, and supervised fine-tuning updates the added memory parameters while preserving the pretrained Qwen weights.

This paper's visual instantiation uses the same PEFT principle with a smaller, explicit registry rather than a large hash vocabulary. Because the personalized concept has a known lexical address, the trigger phrase directly indexes a compact concept table. The rest of the method therefore focuses on where this residual is injected in image and video generators, how its magnitude is calibrated across heterogeneous text encoders, and how it is trained with the frozen visual objective.

\subsection{Visual Reference-Prompt Construction}

The visual training pairs are constructed from a small reference set rather than from a named concept already present in the generator. We use five still images that express the source appearance. To obtain text labels, we apply reverse prompt engineering with GPT-5.3, using its image-input capability to caption each reference into a dense generation-style prompt \citep{openai2026gpt53api}. The prompt describes pose, clothing, accessories, lighting, scene, and rendering style. We then normalize the subject mention: generic phrases such as ``a man'', ``a person'', or ``the character'' are replaced by \trigger{}, while the rest of the prompt is kept as the visual context.

This replacement is important for the experimental control. The prompt should preserve the visual circumstances of each reference image, but the identity slot should be carried only by the fictional trigger. The label construction therefore produces pairs of the form
\[
    (\text{reference image},\ \text{\trigger{} + contextual description}),
\]
rather than prompts that directly name the source character or describe the target identity in ordinary language. For the video extension, we generate short training clips from the same image-prompt pairs using Wan2.2 TI2V in image-to-video mode \citep{wan2025}. The clip labels remain anchored to the normalized source-image prompts so that the image and video objectives share the same identity address. When additional video inspection is needed, sampled frames can be captioned with the same GPT-5.3 reverse-prompting procedure and normalized to the trigger. Thus the video data is a teacher-generated temporal extension of the still-image identity set, not an independently collected video identity dataset.

\subsection{Trigger-Indexed Visual Concept Table}

Each text encoder has its own tokenizer, so the trigger registry is built separately for every encoder. For encoder $m$ with tokenizer $\tau_m$, the trigger is tokenized as
\[
    z^m_{1:T_m} = \tau_m(\trigger{}).
\]
We construct a multi-scale registry from contiguous trigger subspans. For maximum n-gram size $N$, it registers every span $z^m_{i:i+n-1}$ with $2 \leq n \leq \min(T_m,N)$ and also registers the full trigger if $T_m>N$. The reported vision runs use $N=7$. A registry entry is therefore a key-token pair
\[
    k \mapsto z^m_{i:i+n-1},
\]
and each key owns a learned memory vector $e^m_k \in \mathbb{R}^{d_m}$, where $d_m$ is the hidden width of encoder $m$.

At runtime the wrapper scans the tokenized prompt for exact token-id matches against this registry. There is no semantic retrieval, approximate matching, attention-based detector, or prompt rewriting. This exact lookup gives the memory a discrete activation boundary: if none of the registered token sequences appears in the prompt, the Engram table is inactive.

\subsection{Localized Hidden-State Injection}

Let
\[
    H^m = E^{\mathrm{site}}_m(x^m) \in \mathbb{R}^{B \times L_m \times d_m}
\]
be the frozen text states at the conditioning site used by encoder stream $m$ for a batch of tokenized prompts $x^m$. Here the site is the text representation actually consumed by the downstream generator: for example, SD3.5 uses penultimate CLIP token states for its CLIP streams and final T5 encoder states for its T5 stream. For registry key $k$, let $\mathcal{M}^m_k$ be the set of exact matches, with each match represented by a batch index $b$, start position $i$, and span length $n_k$. The Engram residual for key $k$ is added only to positions covered by the matched span:
\[
    \widetilde{H}^m_{b,\ell}
    =
    H^m_{b,\ell}
    +
    \sum_{\substack{(k,i,n_k) \in \mathcal{A}^m_b\\
    i \leq \ell < i+n_k}}
    \delta^m_{b,k},
\]
where $\mathcal{A}^m_b$ is the set of active matches in sample $b$. Because the registry contains multi-scale n-grams, several registered subspans may cover the same trigger token; their residuals are accumulated at that position. When there is no match, the summation is empty, so $\widetilde{H}^m=H^m$.

The simplest residual uses an absolute memory vector with a learned channel scale,
\begin{equation}
    \delta^m_{b,k}=e^m_k \odot s^m .
\label{eq:absolute-residual}
\end{equation}
This form is sufficient for the single-encoder SD1.5 setting. For SD3.5's heterogeneous text encoders, and for Wan2.2's single T5 stream before projection into the video transformer, the reported runs use a bounded per-match relative-norm residual:
\[
    \delta^m_{b,k}
    =
    \operatorname{normalize}(e^m_k)
    \odot
    \tanh(s^m)
    \cdot
    \rho^m_b,
    \qquad
    \rho^m_b =
    \frac{1}{L_m}\sum_{\ell=1}^{L_m}\|H^m_{b,\ell}\|_2 .
\]
Here the memory vector supplies direction, $\tanh(s^m)$ bounds the learned per-channel strength, and $\rho^m_b$ calibrates the update to the activation scale of the current text stream. The motivation differs across backbones. In SD3.5, relative scaling calibrates the Engram update across CLIP-L, OpenCLIP-G, and T5 streams with different widths and hidden-state norms. In Wan2.2, there is only a single T5 stream, but the edited $4096$-dimensional context is later padded and passed through a frozen text projection before video-token cross-attention. Relative scaling therefore acts as an amplitude guard: the injected concept should be large enough to survive the text projection, but not so large that it overwhelms the frozen T5 context. SD1.5 does not require this extra calibration in our implementation because the single CLIP stream is consumed directly by the U-Net cross-attention, so a learned absolute channel scale is tuned against one fixed activation distribution.

\subsection{Adapter-Only Optimization}

Training optimizes only $\{e^m_k\}$ and $\{s^m\}$. All pretrained model components remain frozen, and the original diffusion or flow objective supplies the learning signal. We use separate optimizer parameter groups: concept vectors use the main learning rate, while scale vectors use a smaller learning rate because they directly control residual magnitude. Checkpoints are adapter-only: they store the Engram embeddings and injection scales, while trigger registries and lightweight configuration metadata are either saved with the adapter or reconstructed from the recorded trigger configuration at inference time.

For SD1.5, the frozen VAE encodes each image into latents, a DDPMScheduler adds Gaussian noise, and the frozen U-Net predicts the added noise from the noisy latents and edited CLIP conditioning states. The loss is mean-squared error between predicted and sampled noise:
\begin{equation}
    \mathcal{L}_{\mathrm{SD1.5}}
    =
    \mathbb{E}_{z_0,\epsilon,t}
    \left[
    \|U_\theta(z_t,t,\widetilde{H})-\epsilon\|_2^2
    \right],
    \qquad
    z_t = \operatorname{add\_noise}(z_0,\epsilon,t).
\label{eq:sd15-loss}
\end{equation}

For SD3.5 and Wan2.2, the training scripts use the rectified-flow form implemented by the corresponding frozen transformer. With clean latent $z_0$, Gaussian noise $\epsilon$, and sampled time $t$, the noisy latent is
\[
    z_t = (1-t)z_0 + t\epsilon,
\]
and the model predicts the velocity target $\epsilon-z_0$:
\begin{equation}
    \mathcal{L}_{\mathrm{flow}}
    =
    \mathbb{E}_{z_0,\epsilon,t}
    \left[
    \|F_\theta(z_t,t,\widetilde{C})-(\epsilon-z_0)\|_2^2
    \right],
\label{eq:flow-loss}
\end{equation}
where $\widetilde{C}$ denotes the edited token-level text context after Engram injection. For SD3.5, the frozen pooled CLIP projections are also passed to the transformer as pooled prompt conditioning; they are omitted from the formula because the Engram update is applied to the token context rather than to the pooled projections.

\subsection{Model-Specific Insertion Points}

\paragraph{SD1.5.}
The SD1.5 implementation wraps the single CLIP text encoder used by the latent diffusion pipeline. The CLIP encoder, VAE, and U-Net are frozen. The wrapper injects Engram residuals into CLIP token hidden states before they are consumed by U-Net cross-attention. Because this setting has one text stream with one fixed activation scale, the absolute scaled residual in \cref{eq:absolute-residual} is sufficient in our implementation.

\paragraph{SD3.5.}
The SD3.5 implementation wraps the three text streams used by the Stable Diffusion 3.5 pipeline: CLIP-L, OpenCLIP-G, and T5-XXL. Each stream receives its own tokenizer-specific trigger registry and concept table. For the CLIP streams, the Engram update is applied to the penultimate token states consumed by the SD3 prompt-embedding path; for T5, it is applied to the final encoder states. We follow the SD3 prompt-embedding layout: CLIP-L and OpenCLIP-G token states are concatenated along feature dimension, padded to the T5 width, and concatenated with the T5 sequence along the token dimension,
\[
    [B,77,768] \oplus [B,77,1280]
    \rightarrow [B,77,2048]
    \rightarrow [B,77,4096],
\]
\[
    [B,77,4096] \oplus [B,256,4096]
    \rightarrow [B,333,4096].
\]
The frozen SD3 transformer receives this text sequence as encoder context and receives the frozen pooled CLIP projections as pooled prompt conditioning. The pooled projections are not edited by Engram.

\paragraph{Wan2.2.}
The Wan2.2 extension wraps the single T5 text encoder used by Wan2.2 TI2V-5B. The Wan DiT, Wan VAE, sampler, and T5 backbone are frozen. The wrapper emits an edited variable-length T5 context sequence with width $4096$, which Wan pads to its configured text length and projects into the video-transformer width with its built-in text projection:
\[
    \text{prompt}
    \rightarrow
    \text{frozen T5}
    \rightarrow
    \text{Engram injection}
    \rightarrow
    \text{Wan text projection}
    \rightarrow
    \text{video-token cross-attention}.
\]
Inside Wan2.2, video latents are patchified by a 3D convolution with temporal-spatial patch structure such as $(1,2,2)$ in TI2V-5B. Transformer blocks first apply video self-attention over the spatiotemporal token sequence and then apply cross-attention in which video tokens provide queries and the text context provides keys and values. Thus the Engram memory remains a text-side conditioning edit: video tokens can read the edited T5 context, but the text context is not updated by video tokens. The Wan wrapper averages residuals across overlapping multi-scale matches before returning the edited T5 context, which we use as a small stabilization for the variable-length video text context.

\section{Experiments}

\subsection{Image Personalization Setup}

The image experiments evaluate whether a rare fictional trigger can retrieve a learned visual identity while preserving the surrounding prompt. We use the five reference image-prompt pairs described in the reference-prompt construction above. For both SD1.5 and SD3.5, the trigger is \trigger{}, the registry uses multi-scale n-grams up to length $7$, and all generator and text-encoder weights remain frozen.

For SD1.5, training uses the latent-diffusion noise-prediction objective in \cref{eq:sd15-loss} at $512$ resolution. For SD3.5, training uses the rectified-flow objective in \cref{eq:flow-loss} at SD3 resolution. In both cases, only Engram concept vectors and injection scales are optimized, with separate optimizer parameter groups for embeddings and scales. We compare the frozen base model and the Engram-wrapped model under matched prompts and seeds, and include no-trigger prompts as controls for the activation boundary.

\subsection{Video Data Construction}

The Wan2.2 experiment is an early extension rather than the primary evidence for the method. Because the identity set contains only still images, we first build a teacher-generated video dataset before training the Wan Engram adapter. For each reference image and its normalized trigger prompt, we run frozen Wan2.2 TI2V in image-to-video mode with the source image as visual condition and the trigger-normalized prompt as text condition. The dataset builder saves one generated clip per source image and writes new metadata that maps each video filename back to its training prompt and source image. In the reported run, this produces five $81$-frame training clips at the Wan sampling frame rate, using fixed generation settings for all sources.

\subsection{Video Engram Training}

During Wan2.2 adapter training, each video item is loaded from the generated video dataset, uniformly sampled to nine frames, resized and center-cropped for the Wan VAE, and paired with the metadata prompt. The frozen Wan VAE encodes the frame sequence into video latents, the training loop adds flow-matching noise, and the frozen Wan DiT predicts the velocity target from the noisy video latents and the Engram-edited T5 context. Only the Wan Engram concept vectors and injection scale are optimized. Inference compares frozen-base and Engram-wrapped Wan2.2 generations on a train-style prompt and held-out prompts with matched sampling settings.

\begin{table}[t]
\centering
\small
\begin{tabular}{llll}
\toprule
Backbone & Text stream(s) & Frozen components & Trainable components \\
\midrule
SD1.5 & CLIP & CLIP, VAE, U-Net & Engram vectors, scale \\
SD3.5 & CLIP-L, OpenCLIP-G, T5 & Text encoders, VAE, transformer & Three Engram tables, scales \\
Wan2.2 & T5 & T5, VAE, DiT & T5 Engram table, scale \\
\bottomrule
\end{tabular}
\caption{Trainable components in the visual experiments. All reported visual runs are adapter-only.}
\label{tab:trainable-components}
\end{table}

\section{Results and Discussion}

\subsection{Image Personalization Results}

\begin{figure}[!htbp]
\centering
\includegraphics[width=0.92\linewidth]{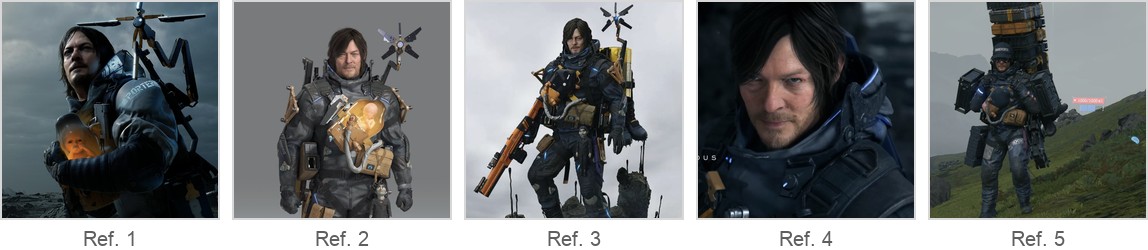}
\caption{Reference images used for visual concept binding. Training prompts are produced by reverse-prompting each image and then replacing the subject mention with the fictional trigger \trigger{}, so the trigger carries the identity slot while the rest of the prompt preserves image-specific context.}
\label{fig:vision-references}
\end{figure}

\begin{figure}[!htbp]
\centering
\includegraphics[width=0.94\linewidth]{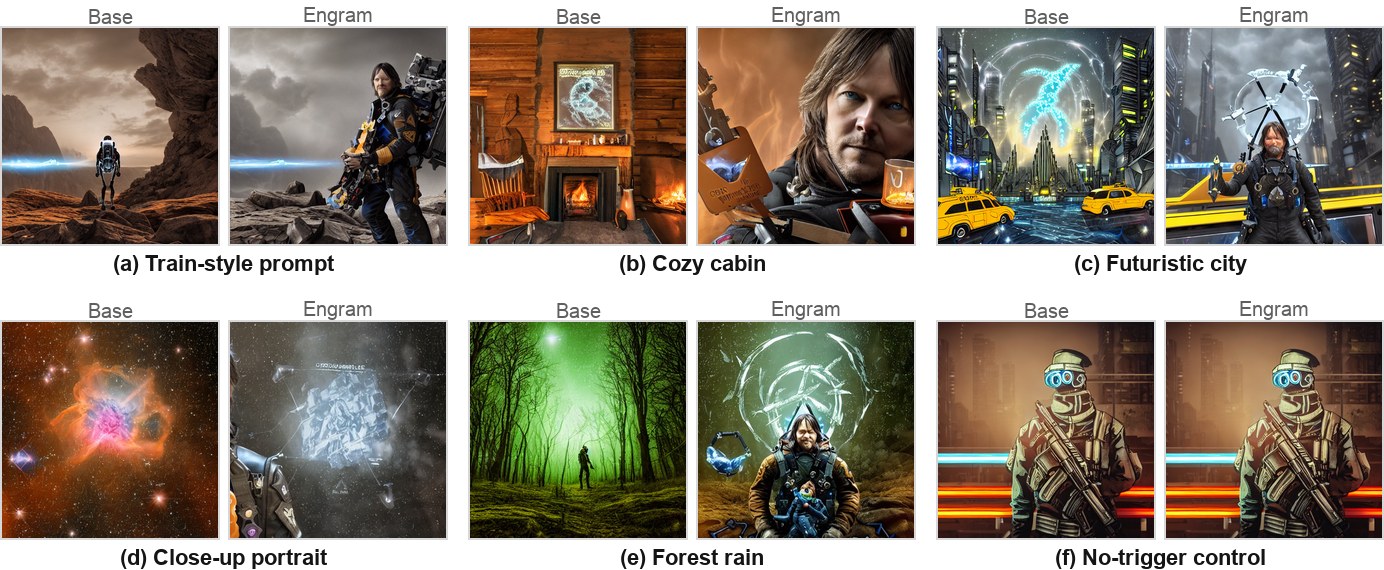}
\caption{SD1.5 qualitative comparison. Each panel pairs the frozen base output on the left with the Engram-wrapped output on the right under the same prompt and seed. Panel titles are shortened display labels; the full prompts are longer and contain \trigger{} in panels (a)--(e). Those panels activate Engram, while panel (f) has no registered trigger span and therefore leaves the Engram branch inactive.}
\label{fig:sd15-results}
\end{figure}

\begin{figure}[!htbp]
\centering
\includegraphics[width=0.94\linewidth]{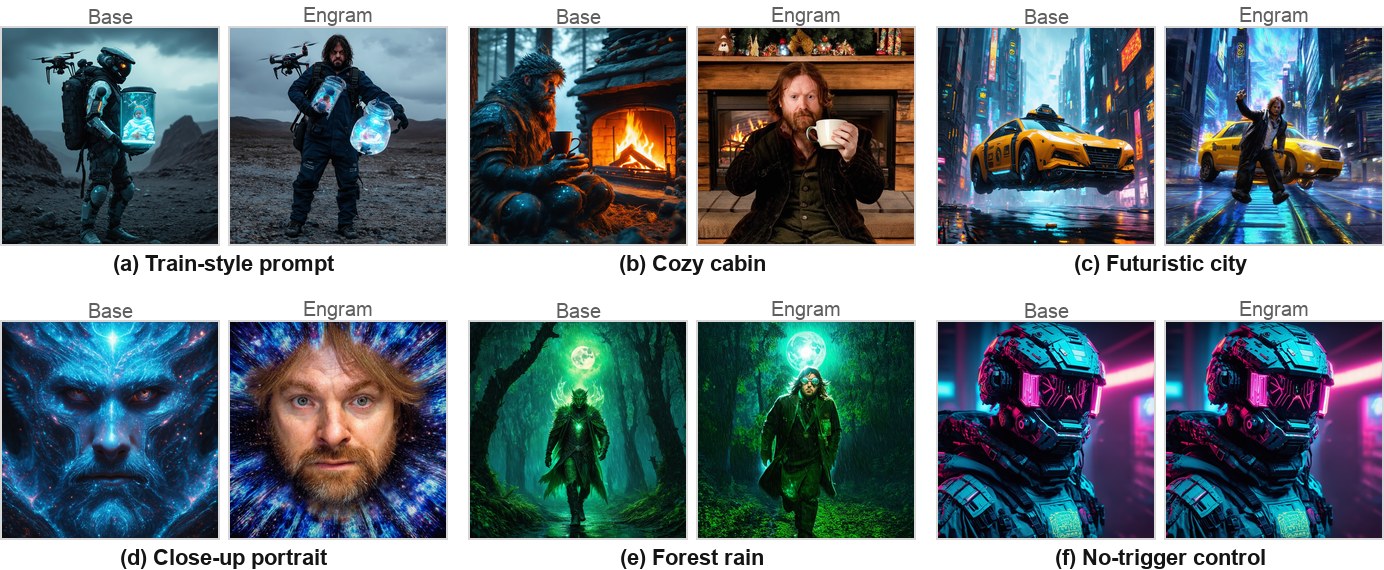}
\caption{SD3.5 qualitative comparison. Each panel pairs the frozen base output on the left with the Engram-wrapped output on the right under the same prompt and seed. Panel titles are shortened display labels; the full prompts are longer and contain \trigger{} in panels (a)--(e). Those panels activate Engram across the CLIP-L, OpenCLIP-G, and T5 token contexts, while panel (f) has no registered trigger span and therefore leaves the Engram branch inactive.}
\label{fig:sd35-results}
\end{figure}

The image experiments test two properties separately: whether the fictional trigger activates a learned visual memory, and whether unmatched prompts remain on the frozen base path. The reference set in \cref{fig:vision-references} contains five still images of the target appearance. The qualitative comparisons in \cref{fig:sd15-results,fig:sd35-results} use matched prompts and seeds. In the base model, \trigger{} is interpreted through its ordinary lexical components and tends to produce generic science-fiction or cybernetic subjects. With Engram enabled, prompts containing the trigger shift toward the reference identity while preserving the surrounding scene, lighting, and composition requested by the prompt.

The SD1.5 result is useful as a minimal single-encoder validation. A single CLIP stream is enough for Engram to override the misleading trigger prior and make the trigger retrieve subject-specific visual features. Its main limitation is inherited from the backbone: prompt adherence, image fidelity, and fine-grained identity precision are less reliable than in the SD3.5 outputs.

SD3.5 gives the stronger image result. The Engram update is distributed across CLIP-L, OpenCLIP-G, and T5 token contexts, and the generated images preserve more of the requested scene while shifting the subject identity. This supports the design choice to inject at the prompt-embedding sites actually consumed by SD3.5 and to use relative-norm scaling across heterogeneous text streams.

The no-trigger controls provide the cleanest evidence for the activation boundary. In both SD1.5 and SD3.5, the saved base and Engram outputs for the control prompt without \trigger{} are byte-identical under matched seeds. This is the expected behavior of the exact n-gram registry: if no registered trigger span is present, no residual is added and the conditioning path is the original frozen path. Thus the image results support a localized personalization claim rather than a globally active style-adapter claim. The image study should still be read as a controlled qualitative personalization result. It does not replace a benchmark comparison against DreamBooth, textual inversion, LoRA, or image-prompt adapters, which would require matched training data, seeds, hyperparameter budgets, and identity metrics.

\subsection{Wan2.2 Video Results}

The Wan2.2 extension is evaluated as a visual activation study rather than as a full video-personalization benchmark. We compare frozen-base Wan2.2 against the same model with an Engram-wrapped T5 encoder. The documented step-28000 comparisons use matched prompts and seeds, $1280\times704$ outputs, $121$ frames, and $24$ fps. In each case, the T5 wrapper is the only intended difference.

\begin{figure}[!htbp]
\centering
\includegraphics[width=0.54\linewidth]{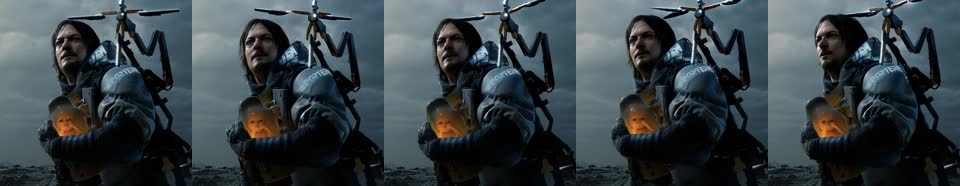}\\[-0.2em]
\includegraphics[width=0.54\linewidth]{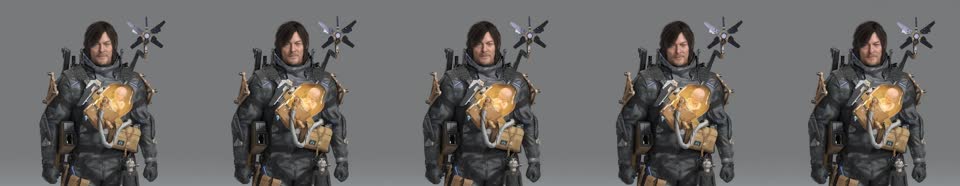}\\[-0.2em]
\includegraphics[width=0.54\linewidth]{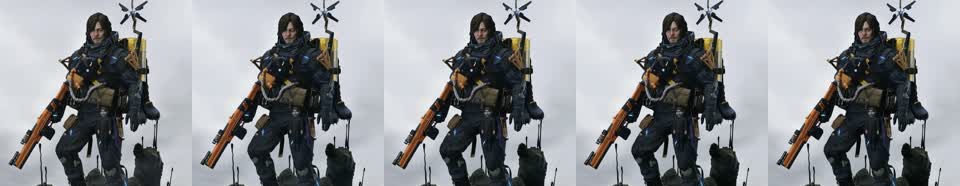}\\[-0.2em]
\includegraphics[width=0.54\linewidth]{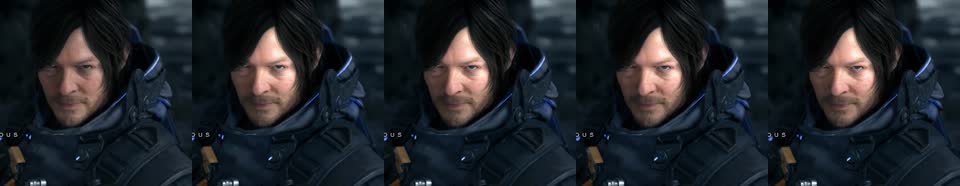}\\[-0.2em]
\includegraphics[width=0.54\linewidth]{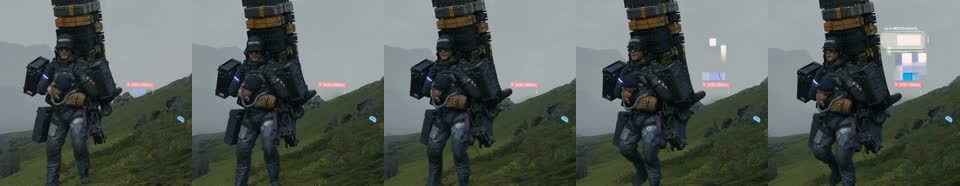}
\caption{Video training clips generated by frozen Wan2.2 TI2V. Each row is a frame contact sheet from one short clip produced using a reference still image and its trigger-normalized prompt. These clips provide the temporal training targets for the video adapter; they are not independently collected identity videos.}
\label{fig:wan-train-clips}
\end{figure}

\Cref{fig:wan-train-clips} shows the video training content. The clips are a temporal extension of the five still-image references: they preserve the same trigger address while adding Wan-generated motion, viewpoints, and scene continuity. This makes the video experiment useful for testing whether the text-side Engram path can influence a frozen video generator, but it also limits the identity claim because the targets inherit the teacher model's own visual bias and temporal drift.

\begin{figure}[!htbp]
\centering
\includegraphics[width=0.96\linewidth]{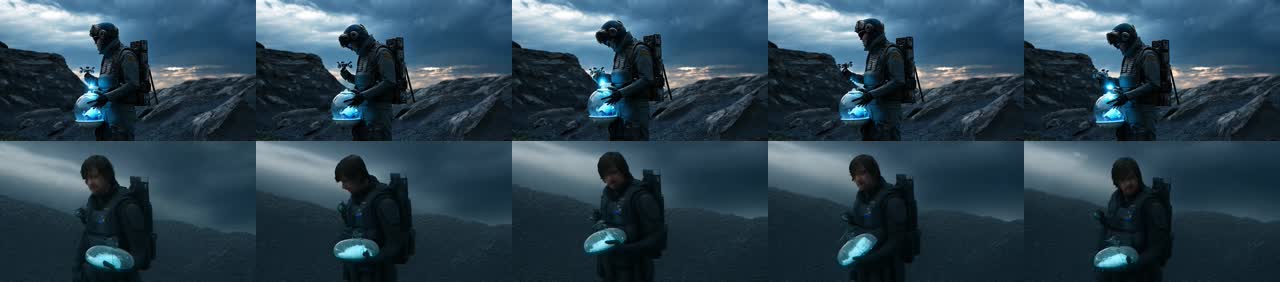}
\caption{Wan2.2 train-prompt comparison at step 28000. The top row shows frozen-base Wan2.2, and the bottom row shows Wan2.2 with the Engram-wrapped T5 encoder under matched prompt, seed, resolution, frame count, and sampling settings. With a train-style prompt containing \trigger{}, the Engram row shifts from a masked generic sci-fi figure toward the long-haired tactical subject encoded by the training clips.}
\label{fig:wan-train-prompt}
\end{figure}

\Cref{fig:wan-train-prompt} gives the clearest Wan2.2 activation result. The trigger match edits the T5 context, and the frozen video stream responds: the Engram output changes subject morphology, costume semantics, and carried-object appearance while retaining the broad scene structure. This supports the narrow claim that the video-side wrapper is active and can steer Wan through the text-conditioning path.

\begin{figure}[!htbp]
\centering
\includegraphics[width=0.96\linewidth]{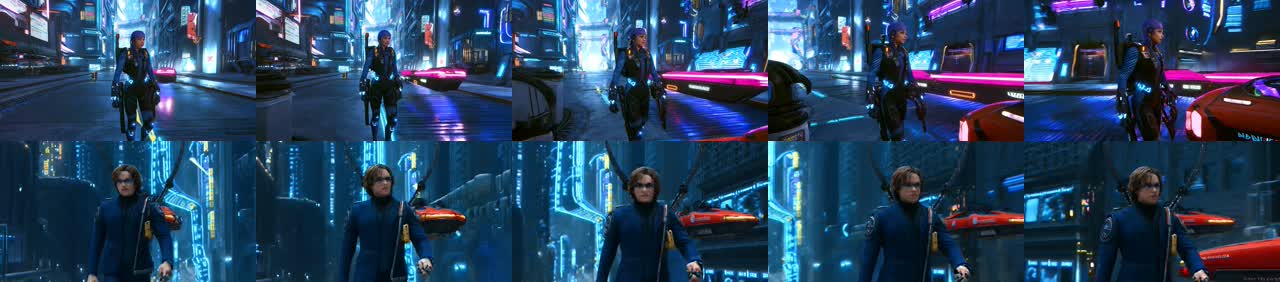}\\[0.2em]
\includegraphics[width=0.96\linewidth]{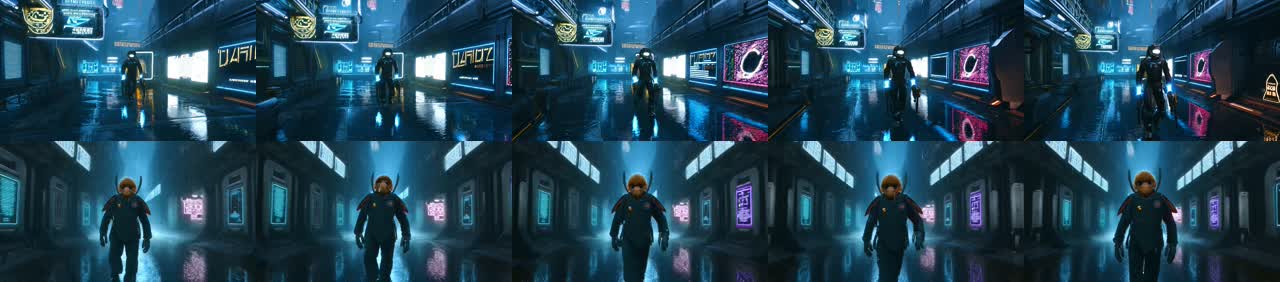}
\caption{Wan2.2 held-out prompt comparisons at step 28000. In each two-row strip, the top row shows frozen-base Wan2.2, and the bottom row shows Wan2.2 with the Engram-wrapped T5 encoder under matched prompt, seed, resolution, frame count, and sampling settings. With held-out prompts containing \trigger{} in new scene contexts, Engram remains visibly active, but the change is weaker and less identity-stable than in the train-style comparison.}
\label{fig:wan-heldout-prompts}
\end{figure}

On held-out prompts, \cref{fig:wan-heldout-prompts} shows a bounded generalization pattern. The Engram branch still changes the generated subject and scene composition under matched settings, but the effect is coarse: it can bias the output toward a long-haired tactical human or alter the subject silhouette, yet it does not reliably reconstruct the fine-grained identity. The current Wan2.2 result is therefore best read as evidence of text-side activation in video generation, not as evidence of fully prompt-composable video identity binding.

\subsection{Discussion}

\begin{figure}[!htbp]
\centering
\includegraphics[width=0.98\linewidth]{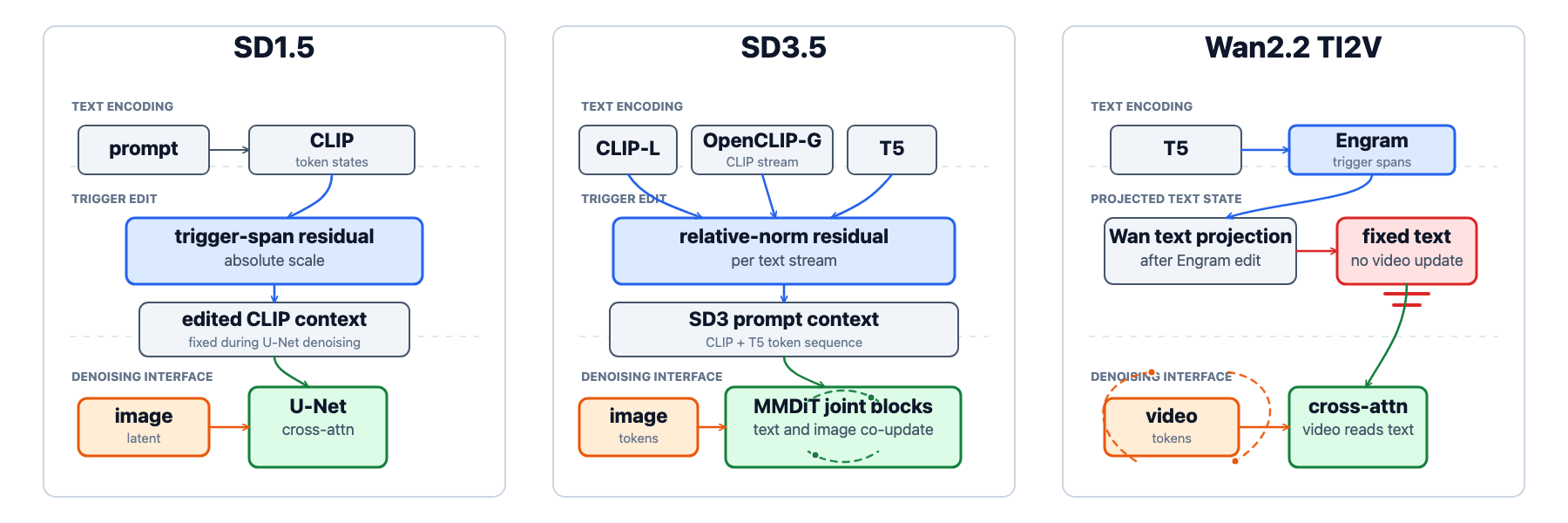}
\caption{Conditioning-path comparison for the three vision backbones. Color key: Engram edit (blue), visual state (orange), attention read (green), fixed text state (red). SD1.5 uses CLIP context read by U-Net cross-attention; SD3.5 feeds edited text tokens into MMDiT joint attention; Wan2.2 exposes a projected T5 context to video-token cross-attention.}
\label{fig:conditioning-comparison}
\end{figure}

The results distinguish activation locality from identity generalization. Activation locality means that a registered trigger changes the matched conditioning states while unmatched prompts follow the frozen path. Identity generalization means that the same trigger continues to denote the target subject under new prompts. The SD image runs support both properties within the evaluated setting. In SD1.5, the residual is added to the CLIP token states consumed by U-Net cross-attention. In SD3.5, the residual is added to the CLIP-L, OpenCLIP-G, and T5 token contexts used to construct the transformer prompt sequence; pooled CLIP conditioning remains frozen. \Cref{fig:conditioning-comparison} summarizes these conditioning paths and highlights why the same text-side Engram edit has different implications across the three backbones.

SD3.5 benefits from this insertion point. Its MMDiT blocks process image and text streams through joint attention, so the edited text tokens participate in the denoising computation rather than serving only as a static side input. The relative-norm rule keeps the residual scale comparable across the three heterogeneous text streams. Under this path, the trigger changes subject identity while the surrounding prompt still controls scene, lighting, and composition. The byte-identical no-trigger controls verify that the effect is governed by the exact n-gram registry rather than by a globally active adapter.

Wan2.2 verifies activation but not precise identity transfer. The train-style comparison shows that the trigger match edits the T5 context and changes the frozen video generator output under matched sampling settings. Held-out prompts also change, but mostly at a coarse semantic level: subject class, silhouette, costume direction, or scene layout. They do not preserve a stable fine-grained identity.

This difference follows from the Wan conditioning path. Wan2.2 maintains an evolving video-token stream that performs spatiotemporal self-attention and then cross-attends to a projected T5 context. Cross-attention updates the video tokens; the T5 context is not updated by video tokens. Engram therefore acts as a fixed text-conditioning modification. This is sufficient to bias semantics, but it provides less direct control over frame-level identity details than an injection path inside a jointly updated text-image transformer.

The video setting also changes the supervision problem. Image personalization optimizes one subject rendering per sample. Video personalization must maintain the same subject across time, pose, camera motion, occlusion, and scene changes. The Wan adapter is trained from only five Wan2.2-generated clips, so the identity signal is narrow and can include teacher drift, motion blur, and inconsistent details. This setup does not directly supervise the broader rule that \trigger{} should denote the same identity across diverse prompts. The present video study should therefore be interpreted as text-side activation with bounded generalization. Improving identity transfer is left to future work; plausible directions include broader prompt coverage, cleaner or filtered video identity targets, and attention-side or block-level conditioning mechanisms closer to the video backbone, in line with prior cross-attention and motion-adapter customization methods \citep{kumari2023custom,ye2023ipadapter,guo2024animatediff}.

\subsection{Limitations}

The evaluation is intentionally small. The image experiments use five reference images and qualitative matched-seed comparisons. The no-trigger controls verify exact activation locality in the saved samples, but we do not yet provide a full identity benchmark or matched comparisons against DreamBooth, textual inversion, LoRA, or image-prompt adapters.

The video extension has stronger threats to validity. The Wan2.2 adapter is trained from only five teacher-generated clips, so the training targets can contain teacher bias, motion blur, and identity drift. The documented step-28000 snapshot contains one train-style comparison and two populated held-out prompt comparisons, all within a sci-fi or cyberpunk prompt family. Future evaluation should add identity embeddings or human identity preference tests, temporal consistency metrics, no-trigger video controls, broader prompt domains, and stronger baselines.

\section{Conclusion}

\method{} implements visual personalization as an explicitly addressed concept table rather than as a globally active adapter. A registered trigger phrase indexes learned memory vectors that modify frozen text-conditioning states only at matched token spans. In the SD1.5 and SD3.5 image settings, this text-side memory provides both activation locality and qualitative identity binding: prompts with the trigger retrieve the target appearance, while prompts without the registered span follow the frozen path.

The Wan2.2 extension gives a narrower result. The same trigger-indexed mechanism can steer a frozen video generator through its T5 context path, but the current setup does not establish prompt-composable or temporally stable identity transfer. This separates text-side activation from video identity generalization. Future work should evaluate broader prompt coverage, cleaner video identity supervision, and conditioning mechanisms closer to the video denoising backbone.

\bibliographystyle{plain}

\end{document}